# Korean to English Translation Using Synchronous TAGs

D. Egedi, M. Palmer, H.S. Park, A.K. Joshi

School of Computer Information Science
and
Institute for Research in Cognitive Science
University of Pennsylvania
Philadelphia, PA 19104-6228, USA

{egedi, mpalmer, hspark, joshi}@linc.cis.upenn.edu

## Abstract

It is often argued that accurate machine translation requires reference to contextual knowledge for the correct treatment of linguistic phenomena such as dropped arguments and accurate lexical selection. One of the historical arguments in favor of the interlingua approach has been that, since it revolves around a deep semantic representation, it is better able to handle the types of linguistic phenomena that are seen as requiring a knowledge-based approach. In this paper we present an alternative approach, exemplified by a prototype system for machine translation of English and Korean which is implemented in Synchronous TAGs. This approach is essentially transfer based, and uses semantic feature unification for accurate lexical selection of polysemous verbs. The same semantic features, when combined with a discourse model which stores previously mentioned entities, can also be used for the recovery of topicalized arguments. In this paper we concentrate on the translation of Korean to English.

## 1 Introduction

It is often argued that accurate machine translation requires reference to contextual knowledge for the correct treatment of linguistic phenomena such as dropped arguments and accurate lexical selection [3, 12]. This is still, in many cases, an unsolved problem for natural language analysis [14], which adds to the burden of the already beleaguered machine translation systems. One of the historical arguments in favor of the interlingua approach has been that, since it revolves around a deep semantic representation, it is better able to handle the types of linguistic phenomena that are seen as requiring a knowledge-based approach. However, recent implementations of machine translation systems are blurring the distinction between transfer systems and interlingua systems [20, 13]. In this paper we present a prototype system for machine translation between English and Korean which is implemented in the Synchronous Tree Adjoining Grammar (STAG) formalism [16], an extension of Lexicalized [15], Feature-Based [19] Tree Adjoining Grammars (FB-LTAGs) [9]. We illustrate the syntactic coverage of the system with relative clauses, wh-questions and constituent order divergence examples. Although we describe what is essentially a transfer based approach, the system uses feature unification for lexical selection and is being augmented with a discourse model to handle discourse related phenomena such as recovery of topicalized arguments. In this paper we concentrate on the translation of Korean to English.

## 2 FB-LTAG and STAG Formalism

### 2.1 An Introduction to FB-LTAGs

The Feature-Based Tree Adjoining Grammar formalism (FB-LTAG) is based on the Tree Adjoining Grammar (TAG) formalism developed by Joshi, Levy, and Takahashi [9], which has been extended to include lexicalization [15], and unification-based feature structures [19]. As first shown by Joshi and Kroch [8, 11], the properties of TAGs permit us to encapsulate diverse syntactic phenomena such as unbounded dependencies in a very natural way.

The primitive elements of the standard TAG formalism are known as **elementary trees**. Elementary trees are of two types: **initial trees** and **auxiliary trees**. In describing natural language, **initial trees** are minimal linguistic structures that contain no recursion, i.e. trees containing the phrasal structure of simple sentences, NPs, PPs, and so forth. Initial trees are characterized by the following: 1) all internal nodes are labeled by non-terminals; 2) all leaf nodes are labeled by terminals or by non-terminal nodes marked for substitution ($\downarrow$).

Recursive structures are represented by auxiliary trees, which represent constituents that are adjuncts to basic structures (e.g. adverbials). **Auxiliary trees** are characterized as follows: 1) all internal nodes are labeled by non-terminals; 2) all leaf nodes are labeled by terminals or by non-terminal nodes marked for substitution, except for exactly one non-terminal node, called the foot node ($*$); 3) the foot node has the same label as the root node of the tree.

Each node of an elementary tree is associated with two feature structures (FS), the top and the bottom. The bottom FS contains information relating to the subtree rooted at the node, and the top FS contains information relating to the supertree at that node. Substitution nodes have only a top FS, while other nodes have both a top and bottom FS.

There are two operations defined in the FB-LTAG formalism, **substitution** and **adjunction**[1]. In the **substitution** operation, illustrated in Figure 1, a node marked for substitution in an elementary tree is replaced by another elementary tree whose root label is the same as the non-terminal. The features of the node at the substitution site are the unified features of the original nodes. The top FS of the node is the result of unification of the top features of the two original nodes, while the bottom FS of the new node is simply the bottom features of the root node of the substituting tree (since the substitution node has no bottom feature).

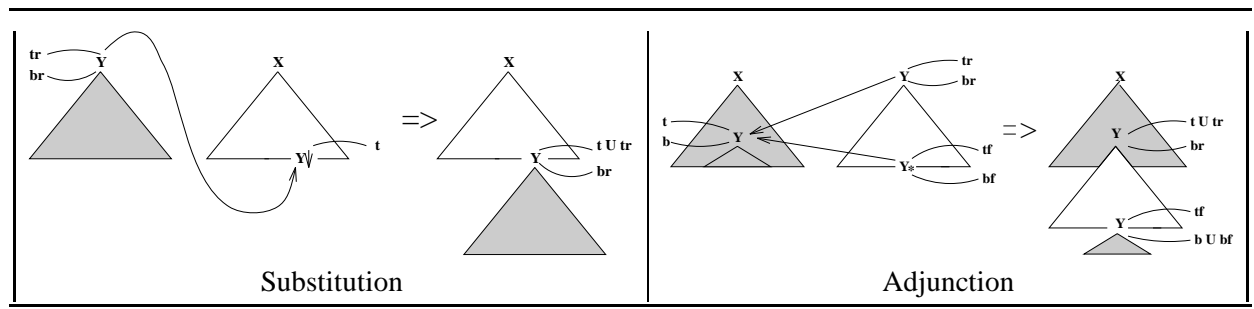

Figure 1: Substitution and Adjunction in FB-LTAG

In an **adjunction** operation, an auxiliary tree is inserted into an initial tree. The root and foot nodes of the auxiliary tree must match the node label at which the auxiliary tree adjoins. The node being adjoined to splits, and its top FS unifies with the top FS of the root node of the auxiliary tree, while its bottom FS unifies with the bottom FS of the foot node of the auxiliary tree. Figure 1 also shows an auxiliary tree and an elementary tree, and the tree resulting from an adjunction operation.

---

[1]Technically, substitution is a specialized version of adjunction, but it is useful to distinguish the two.

## 2.2 An Introduction to Synchronous TAGs

Synchronous Tree Adjoining Grammars (STAGs) are a variant of TAGs introduced by Shieber and Schabes [16] to characterize correspondences between tree adjoining languages. They can be used for relating TAGs for two different languages, or for relating a syntactic TAG and a semantic one for the same language [16, 1] for the purpose of generation [17] or machine translation [2]. STAG has been shown [2, 7] to be capable of handling syntactic and lexical-semantic divergences of the kind shown in Dorr [4].

Transfer rules are stated as correspondences between nodes of the elementary trees of a TAG associated with lexical entries, thus allowing lexical transfer rules to be defined over a large domain of locality. The transfer lexicon specifies a correspondence between lexicalized trees from the source and target grammars. The source sentence is first parsed according to the grammar for the source language. Each elementary tree in the source derivation tree is mapped to a tree in the target derivation tree by looking in the transfer lexicon. These trees are combined according to the links specified between the nodes in the corresponding trees, and the target sentence is read off the final target derivation tree. Correspondences can be made between trees, lexical items, or individual features.

# 3 Translating Korean to English

Korean is an SOV language while English is SVO, so there are some structural differences between the two languages. Figure 2a shows the links between the basic transitive trees for Korean and English.

(1) *ku-ka    ku   pokose-lul   pwunsilhaissta.*
    he-NOM  that  report-ACC   lost-PAST
    *He lost that report.*

Consider translating the Korean sentence in Sentence (1) into English. *pwunsilhaissta* anchors the Korean transitive tree and *lost* anchors the English transitive tree as shown in Figure 2a. The NP trees anchored by the subject and object NP arguments, as well as the determiner trees are shown in Figure 2 as well. Correspondence links are given between each set of Korean and English trees. As the Korean sentence is parsed, the corresponding English structures are pieced together according to the links between the corresponding trees. So, as the subject NP *ku* is substituted into the $NP_0$ node on the transitive Korean tree, the corresponding English NP tree *he* is substituted into the $NP_0$ slot of the English transitive tree, since the linkings between the two transitive trees indicate that those slots correspond. The object NP tree is added in the same fashion, as is the determiner tree. The final translated tree is shown in Figure 3b, and the sentence itself can be simply read from the terminal nodes.

## 3.1 Relative Clauses

Relative clauses in Korean are relatively straight-forward using STAGs. Sentence (2) shows an example sentence in Korean that uses a relative clause. The verb *ssun* (*write*) is in its adnominal form, which indicates that the embedded S is a relative clause. Note that the word being modified, *pokose* (*report*), is on the right side of the relative clause, whereas in English the modified noun is on the left.

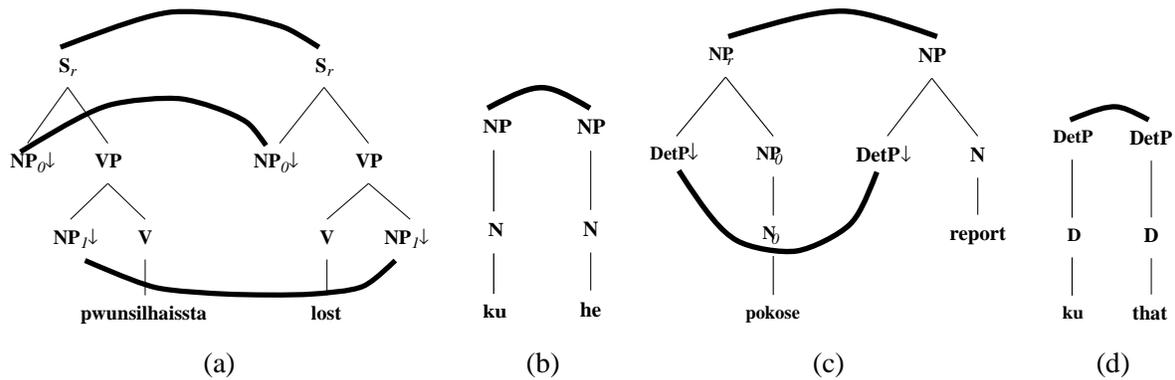

Figure 2: Lexicalized Synchronous trees for *lost*, *he*, *that*, and *report*

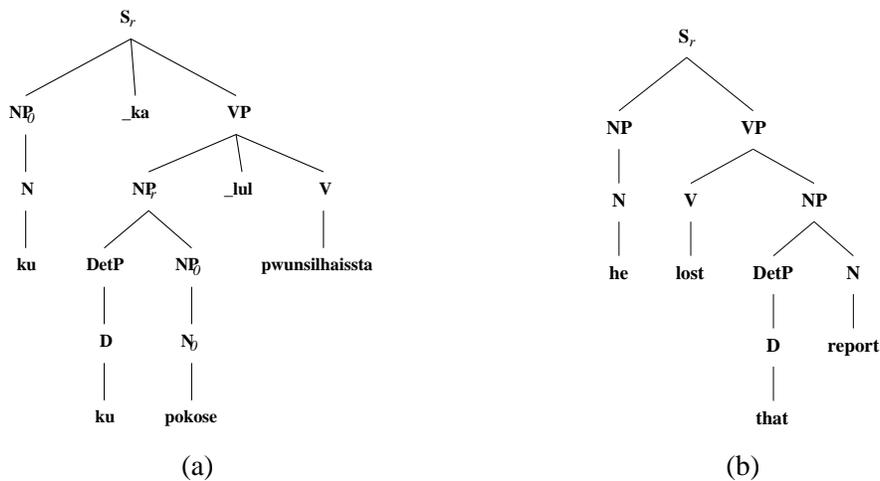

Figure 3: 그가 그 보고서를 분실했다 translated into *He lost that report*.

(2) ku-ka    [kunye-ka ssun    ku   pokose-lul]   pwunsilhaissta.
    he-NOM [she-NOM write-ing that report-ACC] lose-PAST
    *He lost that report that she wrote.*

Since the adnominal form indicates that the embedded S is a relative clause, it will select the Korean relative clause tree. This tree corresponds to the English relative clause tree, which is structurally similar except for the placement of the NP argument on the opposite side of the tree. Since the transfer lexicon specifies mappings between the NP nodes in the two trees, it is a straight-forward process to place the NP argument correctly in the English relative clause tree.

### 3.2 Wh-questions

Translating wh-questions from Korean to English is slightly more interesting than from English to Korean because wh-items do not move in Korean as they do in English. While the basic indicative tree suffices for the wh-question in Korean, a separate structure is required for English. When translating from English to Korean, the English wh-tree simply maps to the Korean indicative tree. However, when translating from Korean to English, the Korean indicative tree must map to one of several possible trees in English, depending on whether the sentence has a wh-word in one of its arguments, and if so, which argument. This highlights one of the strengths of STAGs, since we can take advantage of the feature structures in specifying the mapping. When the mapping of the basic Korean indicative tree to the non-wh English tree is specified, all of the NP nodes in the Korean tree are required to have the feature **[wh–]**. There are separate entries for the Korean trees with the **[wh+]** subject, and a **[wh+]** object, which respectively map to the wh-subject-extraction tree and wh-object-extraction trees for English.

### 3.3 Lexical Selection

The task of lexical selection in machine translation consists of choosing the target lexical item which most closely carries the same meaning as the corresponding item in the source text. In a typical transfer-based approach, a direct mapping is specified from each specific lexical item in one language to a specific lexical item in another. However, there is not always a one-to-one correspondence, and accurate lexical selection can sometimes depend on very subtle semantic distinctions [5, 7] or even require reference to the context [3]. For example, the semantic features that are used in selecting the correct serial verb construction in Chinese (such as the initial shape of the object, choice of instrument) are not all used in selecting English verb senses [21]. The end result is that lexical selection is often predicated on the existence of semantic features that are completely irrelevant to the source language[2].

In Korean, there are several different senses of /WEAR/ that use the same lexical item in English but require distinct Korean expressions, as exemplified in Sentences (3) and (4). In Korean, the verbs for the concept /WEAR/ differentiate between the type of article being worn. English *wear*, however, does not make this distinction. Translating Korean into English is quite simple, as the Korean verbs would each map to English *wear*. However, translating from from English to Korean is more difficult, and we will briefly switch the direction of translation to show how this can be handled.

(3) *ku-ka     yangmal-ul  sinta*
    he-NOM   socks-ACC   wear-PROG
    *He wears socks.*

(4) *ku-ka     os-ul       ipta*
    he-NOM   clothes-ACC wear-PROG
    *He wears clothes.*

We begin with a more coarse-grained lexical translation process, which attempts to focus on a particular set of translation candidates in the source language. These candidates are further narrowed down by a language specific lexical selection process which uses the semantic features associated with the instantiated verb arguments to determine the best fit. Rather than one-to-one mappings between lexical items, the dictionary would map into sets of lexical items.

English *wear* translates ambiguously into all of the Korean /WEAR/ verbs (*sinta*, *ipta*, *ssuta*, *chata*, etc), resulting in a number of trees that correspond to the various verbal projections. Each Korean verb has associated with it a set of selectional restrictions that it imposes on its NP object. For instance *ipta* requires its object to be **[type: bodywear]** and *sinta* requires that its object be **[type: footwear]**, as seen in Trees 4a and 4b. When *yangmal* (*socks*) substitutes into the NP object position, it carries with it its own semantic information, specifically, that it is **[type: footwear]**. When the trees for possible translation have been built and the top and bottom features on each node try to unify, the semantic features on the trees other than the *sinta* tree will clash, and thus fail, as in Figure 4b. The translation for *He wears socks* is *ku-ka yangmal-ul sinta*, as in the tree in Figure 4a.

---

[2]The interlingua approach has a similar difficulty, since it must define an interlingua that can take into account all of the semantic features for both languages. When one begins to consider the problem from the perspective of several languages, this technique quickly become impractical. A vast, language universal ontology must be built that incorporates every semantic feature for every language in an organized fashion.

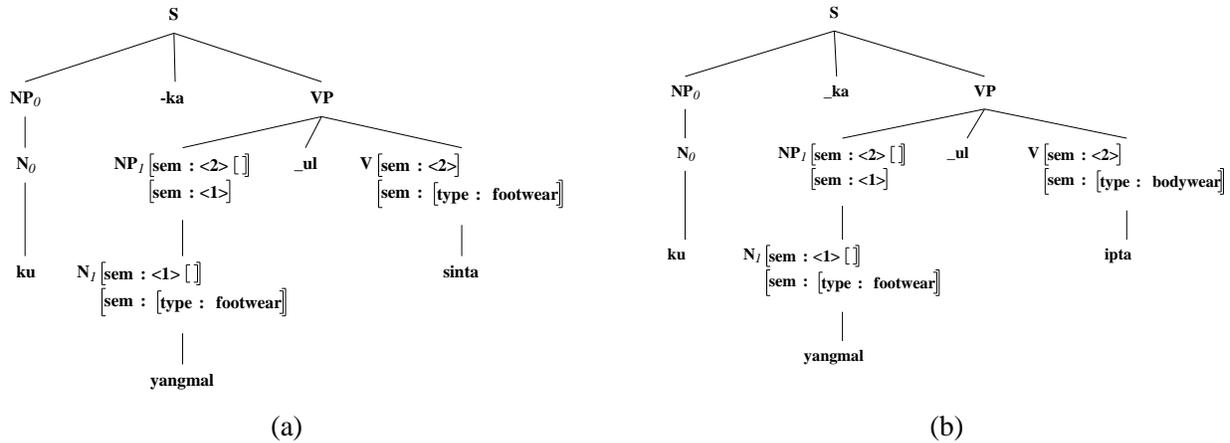

Figure 4: *sinta* and *ipta* trees

### 3.4 NP recovery

One of the most problematic elements in translating Korean text is the widespread use of empty arguments. Korean relies heavily on topic markers, and any argument of a verb can be omitted from a sentence, as long as it can be recovered from the context. The topic marker *-nun* is generally used in Korean to mark new information, and it precedes other information in the sentence. Once mentioned, lexical items that refer to that object may optionally drop as long as they can be understood from context[3]. We present an algorithm here that is aimed at situations where there are two or more potential candidates for a dropped NP position, and one candidate must be selected. We use the feature structures in the STAG formalism, along with ordering rules, to help provide that context. Consider the sentence in (5).

(5)  *Tom-un    pokose-nun  pwunsilhaissta  -ko    malhaissta*
    Tom-TOP  report-TOP   lost            -ko    said
    *Tom said he lost the report.*

The two noun phrases in the sentence: *Tom* and *pokose* are both marked with the topic marker *-nun*[4]. The parse of this sentence has empty arguments for the two verbs, with the topic markers adjoined on[5].

---

[3]There are several contextual factors, both pragmatic and semantic, involved in insuring that there is an unambiguous reference for a dropped NP. Obviously, inherent lexical semantic constraints on argument filling come into play. For instance, the sentence *Mary said John threatened her* is not ambiguous, because John cannot threaten himself. In such a case, our algorithm would need not apply.

[4]Because of phonological considerations, the *-nun* marker becomes *-un* after a consonant.

[5]The adjunction of the topic markers is optional, since they can be dropped completely if the noun phrase is recoverable from the wider context.

To recover the missing noun phrases, we use an ordered list that contains information about noun phrases previously mentioned in the discourse. This is similar to the techniques used in recovering elided noun phrases in English [14] and dropped subjects in Japanese [10], and for resolving English anaphora [6, 18]. For each dropped argument, we look through the list, matching the semantic feature constraints from the verb against the features on the NP until an appropriate NP is found.

For example, in Sentence (5) each topic is added to the ordered list along with its associated semantic features. After processing the topic NPs, the list would be {**pokose [animate–]**, **Tom [animate+]**}. *malhaissta* (*said*) imposes a **[animate+]** semantic feature constraint on its subject. Looking at the beginning of the ordered list, we see that the closest topic is *pokose* (*report*), which is **[animate–]**, so it can not fill that argument. The next topic is *Tom*, which does meet the restrictions on the argument, so it fills the subject argument slot. The next empty argument is the subject of the embedded clause, which we will skip over since we believe that it is actually an instance of PRO-control[6]. The next empty argument is the object of the embedded verb ($NP_2$) *pwunsilhaissta* (*lost*), which is constrained to be **[animate–]**. Again, the closest topic is *pokose* (*report*), which fits the semantic constraints, and so fills the object argument slot. Once the arguments have been recovered, the augmented Korean sentence can be translated using the STAG formalism, as illustrated earlier.

If the sentence had been in an appropriate context, then it would have been possible for one or both of the topic NPs to have been completely missing from the sentence. In this case, the information to recover the missing NPs would have to come from a global ordered list, as well as a local one.

## 4  Conclusion

We have presented a prototype Synchronous TAGs translation system, augmented with semantic feature unification. This system is being applied to a domain of military messages for translation between Korean and English. To illustrate the coverage of the system and the effectiveness of semantic feature unification when combined with a simple discourse model, we have given examples of various syntactic phenomena, e.g., relative clauses and wh-questions; semantic phenomena, e.g., accurate lexical selection for polysemous verbs; and pragmatic phenomena, e.g., the recovery of topicalized arguments.

---

[6]There are strong arguments for the subject of the embedded clause to be an instance of PRO-control and as such it would get its reference from the subject of the matrix clause. We do not want to delve into those arguments here, as it is certainly possible for the subject of the embedded clause to undergo the argument recovery process as well, if one wanted to argue against a PRO control analysis.